# Processing South Asian Languages Written in the Latin Script: the Dakshina Dataset

**Brian Roark†, Lawrence Wolf-Sonkin†, Christo Kirov†, Sabrina J. Mielke‡,
Cibu Johny†, Işın Demirşahin† and Keith Hall†**

†Google Research
{roark,wolfsonkin,ckirov,cibu,isin,kbhall}@google.com

‡Johns Hopkins University
sjmielke@jhu.edu

**Abstract**
This paper describes the Dakshina dataset, a new resource consisting of text in both the Latin and native scripts for 12 South Asian languages. The dataset includes, for each language: 1) native script Wikipedia text; 2) a romanization lexicon; and 3) full sentence parallel data in both a native script of the language and the basic Latin alphabet. We document the methods used for preparation and selection of the Wikipedia text in each language; collection of attested romanizations for sampled lexicons; and manual romanization of held-out sentences from the native script collections. We additionally provide baseline results on several tasks made possible by the dataset, including single word transliteration, full sentence transliteration, and language modeling of native script and romanized text.

## 1. Introduction

Languages in South Asia – a region covering India, Pakistan, Bangladesh and neighboring countries – are generally written with Brahmic or Perso-Arabic scripts, but are also often written in the Latin script, most notably for informal communication such as within SMS messages, WhatsApp, or social media. While the use of the Latin script as the means for inputting text (romanization) is relatively common for other languages, South Asian languages lack common standard romanization systems (such as pinyin for Chinese). In other words, when individuals use the Latin script to write in South Asian languages, they do not adhere to a system, i.e., there is no commonly used standard Latin script orthography in these languages. Rather, individuals generally use the Latin script to provide a rough phonetic transcription of the intended word, which can vary from individual to individual due to any number of factors, including regional or dialectal differences in pronunciations, differing conventions of transcription, or simple idiosyncrasy. For example, Wolf-Sonkin et al. (2019) present an example from the comment thread to a blog entry, which includes Hindi comments in both the native script Devanagari and in the Latin script, where the word भ्रष्टाचार is romanized variously as *bhrastachar*, *bhrashtachar*, *barashtachaar* and *bharastachar*, among others.

The prevalence of the Latin script for writing these languages has led to, among other things, work on text-entry systems for those languages in the Latin script, including systems that transliterate from Latin script input to the native script of the language (Hellsten et al., 2017), as well as systems that produce text in the Latin script but provide model-based features such as auto-correction (Wolf-Sonkin et al., 2019). Note, however, that this sort of script variance, beyond just complicating systems for text entry, is also a challenge for common natural language processing (NLP) applications such as information retrieval and extraction, machine translation, and written dialogue interactions. Many important component NLP tasks – such as language identification – are also impacted by such text. Further, for historical reasons and due to the role of English as a regional lingua franca, the prevalence of loanwords and code switching (largely but not exclusively with English) is high.

Unlike translations, human generated parallel versions of text in the native scripts of these languages and romanized versions do not spontaneously occur in any meaningful volume. Further, given the generally informal nature of Latin script text in South Asian languages, much of the content is interpersonal communication, hence not generally available to serve as training data for approaches dealing with this phenomenon. While data can be found on the internet, such scrapings are of mixed provenance and are unlikely to have straightforward data licensing. Further, finding such data likely would rely on language identification algorithms that themselves require training.

For these reasons, we are releasing the Dakshina dataset[1], consisting of native script text, a romanization lexicon and some romanized full sentences,[2] all derived from Wikipedia data in 12 South Asian languages. This data can be used for validation on a range of real-world use scenarios, including single-word or full-sentence transliteration from the Latin script back to the native script, or language modeling in either the native script or Latin script. In this paper, we provide a number of baselines for these tasks, including both neural and finite-state approaches, in addition to full documentation of the dataset creation.

## 2. Background

### 2.1. Scripts in South Asian languages

Brahmic scripts are widely used throughout South Asia. They are abugida (also known as alphasyllabary) scripts where consonant characters come with an inherent (or default) vowel. For example, the character स in the Devanagari script (used by both Hindi and Marathi in our dataset) represents the akṣara (or orthographic syllable) 'sa'. If another vowel, or no vowel, is needed for the consonant, then either a pure consonantal form or a ligature with the vowel sign is used. Thus, स् uses the virama diacritic to represent the consonant 's' with no vowel, and सं represents a nasalized vowel, i.e., 'san'. These glyphs are represented as strings of Unicode characters that are then rendered into the glyph, e.g., सं is a pair of Unicode characters: स ं. When multiple consonants are used in sequence, they are combined into a single ligature, e.g., 'skri' from the word संस्कृत (Sanskrit) is written स्कृ. This ligature is actually represented by a string of four Unicode characters: स ् क ृ. Native script keyboards in these scripts are less common than Latin script input methods – similar to Chinese. Unlike methods such as pinyin in Chinese, however, there are generally no standard romanization systems in common use for the languages using Brahmic scripts. Rather, as pointed out in Wolf-Sonkin et al. (2019), individuals use the basic Latin script to provide a rough phonetic transcription of the words in these languages. Interestingly, languages from the region using the Perso-Arabic script – such as Sindhi and Urdu in this collection – also widely use romanization for text input, despite the relative ease of representing this consonantal writing system in a native keyboard.

---

[1] https://github.com/google-research-datasets/dakshina

[2] We use the term 'sentence' here to contrast to isolated words, even though some multi-word strings may not be full sentences.

That the romanizations are largely driven by spoken pronunciation rather than the written forms is evidenced by the fact that the same native script character may be romanized differently depending on coarticulation effects, e.g., in Hindi सं is typically romanized as 'sam' when the following akṣara begins with labial closure (संपूर्ण, 'sampurn') versus not (संस्कृत, 'sanskrit'). Also of note is that those romanizing South Asian languages from Perso-Arabic scripts romanize with vowels, despite not typically being included in the native writing system.

## 2.2. Transliteration & Romanized text processing

Early NLP work on automatic transliteration between writing systems was driven by the needs of machine translation or information retrieval systems, and hence was generally focused on proper names and/or loanwords (Knight and Graehl, 1998; Chen et al., 1998; Virga and Khudanpur, 2003; Li et al., 2004). Pronunciation modeling did play a role in early approaches (Knight and Graehl, 1998), though directly modeling script-to-script correspondences was eventually shown to be effective (Li et al., 2004). Advances have been made in transliteration modeling in a range of scenarios – see Wolf-Sonkin et al. (2019) for further citations – including the use of transliteration in increasingly challenging settings, such as information retrieval with mixed scripts (Gupta et al., 2014) or for text entry on mobile devices (Hellsten et al., 2017).

As SMS messaging and social media have become more and more ubiquitous, Latin script input method editors (IMEs) have become increasingly common for languages with other native scripts, leading to increasing amounts of romanized text in these languages (Ahmed et al., 2011). The spelling variation that arises from the above-mentioned lack of standard orthography in the Latin script in South Asian languages is similar to issues found in the writing of dialects of Arabic (Habash et al., 2012), and in the processing (e.g., optical character recognition) of historical documents from prior to spelling normalization (Garrette and Alpert-Abrams, 2016). In fact, dialectal Arabic is also frequently romanized, compounding the problem by lacking a standard orthography on either side of transliteration (Al-Badrashiny et al., 2014). Automatic processing of romanized text for languages using Perso-Arabic scripts is relatively common in the NLP literature, including, of course, Arabic – e.g., the above citations and Chalabi and Gerges (2012) – but also Persian (Maleki and Ahrenberg, 2008) and Urdu (Irvine et al., 2012; Bögel, 2012; Rafae et al., 2015).

Romanization can make language identification particularly difficult, as multiple languages may share similarities in their romanized strings. In fact, Bögel (2012) used rule-based XFST-encoded transliteration between romanized Urdu and its native Perso-Arabic script in both directions, to allow the romanized text to be used as an intermediate representation between Hindi and Urdu. In addition, code switching is common, e.g., between Japanese and English in language learners (Nagata et al., 2008), between dialectal and modern standard Arabic (Eskander et al., 2014), or between dialectal Arabic, French and English (Voss et al., 2014). Adouane et al. (2016) looked at a language identification scenario that included both romanized dialectal Arabic and romanized Berber. Zhang et al. (2018) looked at large-scale language identification that included some evaluation of Hindi written in the Latin script.

While most research looks at transforming romanized text into a native script, some has looked at automatic romanization, as the means of, for example, normalizing SMS messages or other texts into some formal romanization standard, such as those used for glossing of linguistic texts (Aroonmanakun, 2004; Kim and Shin, 2013). Others have simply tried to perform text normalization directly, e.g., based on clustering (Rafae et al., 2015).

Language modeling of romanized text for these languages, with their lack of orthography, has been relatively under-explored. There are important use scenarios for such models, including the potential for language identification of romanized text – a task that has been studied though without the benefit of advanced language modeling methods – but also for models used in mobile keyboards for auto-correction, gesture-based text entry and/or word prediction and completion (Wolf-Sonkin et al., 2019).

The lack of standard orthography in the Latin script in these South Asian languages, coupled with the fact that large repositories of parallel text in these scripts do not arise spontaneously, makes modeling (and validating models) challenging. The Dakshina dataset is intended to provide publicly available data for training and validating models in a diversity of regional languages.

## 3. Data

The data in the Dakshina dataset is all based on Wikipedia text, extracted in March, 2019, in 12 South Asian languages. The collected languages, and statistics about the quantity of data per language are shown in Table 1. Four of the 12 languages are Dravidian languages (kn, ml, ta, te), while the rest are Indo-Aryan languages. The text in two of the languages (sd, ur) is written in Perso-Arabic scripts; the rest

| Language | BCP 47 tag | Script | Native script Wikipedia | | | Romanization Lexicon | | | Romanized Wikipedia | |
|---|---|---|---|---|---|---|---|---|---|---|
| | | | sentences (x1000) | | words per sentence | types (native script) | entries dev/test | entries per type | sentences romanized | native words per sentence |
| | | | training | validation | | train | | | | |
| Bengali | bn | Bengali | 895.1 | 26.7 | 12.2 | 25,000 | 5,000 | 3.8 | 10,000 | 12.1 |
| Gujarati | gu | Gujarati | 146.1 | 24.8 | 15.3 | 25,000 | 5,000 | 4.2 | 10,000 | 15.4 |
| Hindi | hi | Devanagari | 1065.0 | 24.9 | 17.3 | 25,000 | 5,000 | 1.8 | 10,000 | 17.7 |
| Kannada | kn | Kannada | 678.9 | 23.6 | 11.0 | 25,000 | 5,000 | 2.0 | 10,000 | 11.1 |
| Malayalam | ml | Malayalam | 747.5 | 25.8 | 8.7 | 25,000 | 5,000 | 2.3 | 10,000 | 8.6 |
| Marathi | mr | Devanagari | 361.3 | 25.8 | 10.1 | 25,000 | 5,000 | 2.3 | 10,000 | 9.8 |
| Punjabi | pa | Gurmukhi | 242.3 | 26.1 | 17.5 | 25,000 | 5,000 | 2.8 | 10,000 | 17.8 |
| Sindhi | sd | Perso-Arabic | 77.9 | 28.0 | 19.0 | 15,000 | 5,000 | 2.6 | 9,999 | 18.5 |
| Sinhala | si | Sinhala | 200.6 | 28.6 | 14.2 | 25,000 | 5,000 | 1.7 | 10,000 | 14.3 |
| Tamil | ta | Tamil | 1144.4 | 26.2 | 9.6 | 25,000 | 5,000 | 2.7 | 10,000 | 9.5 |
| Telugu | te | Telugu | 874.4 | 24.7 | 9.9 | 25,000 | 5,000 | 2.3 | 10,000 | 9.4 |
| Urdu | ur | Perso-Arabic | 507.3 | 25.7 | 17.2 | 25,000 | 5,000 | 4.2 | 9,759 | 17.4 |

Table 1: Data quantity by language.

are written in Brahmic scripts.[3]

For each language, there are three types of data. First, there is native script Wikipedia text in the language, which is split into training and validation partitions. Information about pre-processing of the text, and what raw data is included in the collection is provided in §3.1.. Second, there is a romanization lexicon in each language, akin to a pronunciation lexicon such as Weide (1998), whereby words in the native script are accompanied by some number of attested romanizations. §3.2. will detail the process by which these annotations were derived. Finally, 10,000 sentences from the validation partition of the native script text detailed in §3.1. were manually romanized by annotators – see §3.3. for details on the collection of these romanizations. In addition, the README at the Dakshina dataset URL provided in footnote 1 contains some additional details on the dataset beyond what is covered here.

### 3.1. Native script text

Our intent in collecting native script text from Wikipedia was to provide a useful extraction (and partitioning into training and validation sets) of non-boilerplate full sentences that fall primarily in the Unicode block of the native script. We collected the text in a way that allows for relatively easy tracking of each sentence back to the specific Wikipedia page where it occurred. This provides some degree of future flexibility, e.g., taking into account considerations that we are not currently tracking, such as topic; or perhaps using some other kind of sentence segmentation. We perform minimal pre-processing (detailed below) and provide intermediate data where useful. Finally, the data is partitioned into training and validation by Wikipedia page, so that sentences in the validation sets never come from the same Wikipedia page as any sentences in the training set.

The first task of extraction was to ensure that we were extracting non-boilerplate sentences primarily in the Unicode block of a native script of the language. We were not interested in, for example, extracting tabular data, lists of references, or boilerplate factoids about locales, all of which occur with some frequency in Wikipedia. Through some amount of trial-and-error, we determined a set of high-level criteria for excluding Wikipedia pages that consist largely of such material from our collection. We chose to pursue such a strategy, rather than, say, relying on some kind of modeling solution (e.g., language models) to avoid biasing the data with assumptions implicit in such models. We are explicit about the criteria that were used to exclude pages, and the identities of omitted pages are included with the corpus. We omitted pages:

- that redirect to other pages.
- containing infoboxes about settlements or jurisdictions.
- containing templates with a parameter named `state` valued either `collapsed`, `expanded` or `autocollapse`.
- referring to censusindia or en.wikipedia.org.
- with wikitable or lists containing more than 7 items.

From pages that were not excluded from the collection by the above criteria, we then aggregated statistics by section heading, to determine sections that generally consist of a significant amount of text outside of the block of the native script. For example, reference sections tend to have a very high percentage of Latin script relative to other sections in the Wikipedia pages of these South Asian languages.

Determining the amount of text in collected sentences to allow from outside of the native script Unicode block can be tricky, because such characters can occur naturally in fluent text, e.g., Basic Latin block punctuation and numerals, or parentheticals giving the Latin script realization of a proper name. For this reason, for each native script, we defined two sets of characters: special non-letter characters $\mathcal{N}$, and characters within the native script Unicode block $\mathcal{B}$. The special non-letter characters $\mathcal{N}$ come from across Unicode, including non-letter (i.e., not a-z or A-Z) Basic Latin block characters; Arabic full stop (U+06D4); Devanagari Danda (U+0964); any character in the General Punctuation block; and any digits in $\mathcal{B}$. Note that the intersection $\mathcal{N} \bigcap \mathcal{B}$ is generally non-empty.

For individual sentences or whole sections of text, we can aggregate statistics about how many characters fall within $\mathcal{N}$ and $\mathcal{B}$, as well as how many fall within their intersection. Our exclusion criteria are based on three values:

---

[3]The Wikipedia pages that we accessed for Punjabi were in the Brahmic Gurmukhi script not the Perso-Arabic Shahmukhi script.

1. The fraction of characters $c$ from a string or section such that $c \notin \mathcal{N}$ and $c \notin \mathcal{B}$. (Should be low.)

2. The fraction of characters $c$ from a string or section such that $c \in \mathcal{B}$. (Should be high.)

3. The fraction of whitespace-delimited words from a string or section with at least one character $c$ such that $c \in \mathcal{B}$ and $c \notin \mathcal{N}$. (Should be high.)

We first aggregate statistics about sections with a specific title across all pages, and if the percentage of characters $c$ from all sections with that title such that $c \notin \mathcal{B}$ and $c \notin \mathcal{N}$ is above 20%, then sections with that title are omitted. For example, the section title सन्दर्भ in the Hindi collection has over 50% of its characters falling outside of $\mathcal{B}$ (Devanagari in this case) and not in $\mathcal{N}$, which is perhaps not surprising for a section title that translates as 'references'.

After filtering sections identified as above, we then segment sections into individual sentences. We do not train segmentation models, rather rely on simple deterministic methods, to avoid biasing the collection. First, we use any newline segmentation present in the Wikipedia data to segment sections. Next we use the ICU[4] sentence segmenter, initialized with the locale for the language being processed. Additionally, we perform NFC normalization on all sentences in the collection. This leaves us with segmented sentences, which are then also filtered based on our three values enumerated above. In particular, we include only sentences that have: (1) at most 10% characters $c$ such that $c \notin \mathcal{B}$ and $c \notin \mathcal{N}$; (2) at least 85% characters $c$ such that $c \in \mathcal{B}$; and (3) at least 85% of whitespace-delimited words have at least one character $c$ such that $c \in \mathcal{B}$ and $c \notin \mathcal{N}$.

As with earlier filtering stages, we include details in the dataset of the sections and sentences that are omitted from the collection, along with the origin page information for all sentences, to allow for other methods of filtering or pre-processing by those making use of the dataset.

As can be seen from Table 1, the number of sentences and the words per sentence does vary, from less than 100 thousand sentences in Sindhi to over a million in Hindi and Tamil; and from less than 10 words per sentence in Dravidian languages Malayalam, Tamil and Telugu; to between 17 and 19 words for Hindi, Punjabi, Sindhi and Urdu.

### 3.2. Romanization lexicons

For each of the languages in the dataset, we additionally provide a romanization lexicon, consisting of entries pairing native script words with romanizations attested by native speaker annotators. We also give the number of times each romanization was attested during data collection. Note that this is not a frequency of usage, but is rather the number of times the romanization was elicited from an annotator for the given native form. The romanizations were elicited through an initial request and a follow up elicitation. For each native script word, first, they were asked, "How would you write this word in the Latin script?" After providing their romanization of the word, they were asked, "Can you think of any other ways that the word is written in the Latin script?" The third column of the each lexicon is the number of attestations. Thus, for example, entries for the Tamil

---

[4] http://icu-project.org

word அகதிகள் (refugees) in the training lexicon are:

    அகதிகள்    agathigal    2
    அகதிகள்    akathigal    1

indicating that the first romanization is attested twice, while the second one was attested just once.

It should be noted that a substantial number of words in many of the lexicons are English loanwords, which present something of a special case, to the extent that the standard English orthography is typically the most commonly attested romanization of the word. This may be accompanied by romanizations that diverge from English orthography, particularly since English spellings are relatively opaque to pronunciation. For example, in the Tamil training lexicon:

    டெம்பிள்    tempil    1
    டெம்பிள்    temple    3

the English word "temple" receives 3 attestations of the correct spelling alongside one attestation of a phonetically spelled romanization. Different annotators may choose to attest more alternatives than others, and for some languages we end up with many alternatives for certain exemplars, as evidenced by the "entries per type" column in Table 1.

The lexicon in each language consists of a sampling of native script words that have frequency greater than one in the training section of the Wikipedia text for that language, as described in § 3.1.. We have 30,000 entries for all languages except Sindi, where we have 20,000. For convenience, we have partitioned the data, with 5,000 being considered validation examples (split evenly between development and test sections), and the rest in a training section.

To improve the training/validation separation, we ensured that no validation set word shares a lemma with any training set word. To obtain lemmata (of which there may be multiple) for each word, we used Morfessor FlatCat (Grönroos et al., 2014) to train a model for each language on the first 77853 lines[5] of its shuffled Wikipedia text. Using standard hyperparameters yields slightly over-segmented (and thus more conservatively separated) lemmata for each word in training and validation sets.

### 3.3. Romanized Wikipedia

Finally, we provide romanizations of full Wikipedia sentences, randomly selected from the validation set of the native script Wikipedia collections described in § 3.1.. For each language in the dataset, we had 10,000 sentences romanized by native speakers. We split long sentences (greater than 30 whitespace delimited tokens) into shorter strings for ease of annotation, recursively partitioning each string at the halfway point until strings below the length threshold were obtained. After annotation, these segmented strings were rejoined, and the resulting pairs of native script and romanized sentence pairs are provided in the dataset in both their split and rejoined versions.

Annotators were instructed to write the given sentences as they would if they were writing them in the Latin script. Strings and characters already in the Latin script, or outside of the native script Unicode block, were to be passed through to the romanized sentence unchanged. Beyond this, we provided no specific guidance on romanization strategy.

---

[5] This is the most text available in all languages, allowing comparability when manually inspecting segmentation granularity.

In Urdu and Sindhi, a small number of sentences (just over 200 in Urdu and just one in Sindhi) ended up being from another language written in the Perso-Arabic script, and these sentences were omitted from the set, as can be seen from the number of sentences in Table 1. Again, there was no filtering or biasing of the data based on existing language identification models; rather filtering was based on script and page or section characteristics, leading to this small number of examples from outside the language.

We treat these 10,000 sentences as a validation set, and split each language in half into dev and test sets. For example, from the tab delimited Hindi dev set:

जबकि यह जैनों से कम है।    Jabki yah Jainon se km hai.

A single romanization was elicited for each sentence. These resulting romanizations were then put through a validation phase, where they were sent to annotators who were asked to transcribe them in the native script of the language. This round-trip validation cannot achieve perfect accuracy, due to a few considerations. First, some sentences have a certain amount of content outside of the native script Unicode block, as described in § 3.1.. Second, there is some variability when it comes to writing: things like punctuation or digits can come from the native block – e.g., Danda for full stop as in the example above, or basic Latin block period (.); or the Devanagari digit ५ and the Latin script digit 5, which represent the same value. We see the representation of years, for example, in both the native and Latin script digits in the native script Wikipedia,[6] something that cannot be determined when just given the romanized version. Additionally, there is some variability of spelling in the native scripts, e.g., of proper names or English loanwords.

The results of this validation phase are also released in the dataset, along with an assessment of the match between each original native script string and the string produced during validation. To perform this assessment, we put both the original strings and validation output through extensive additional visual normalization beyond NFC normalization, whereby visually identical strings were mapped to the same string of Unicode characters. While this visual normalization does not capture all equivalences, it does reduce spurious mismatches to some degree. Mean character error rate over the set between the original and validation strings after normalization was 0.054 (standard deviation 0.012), with a high of 0.084 (Sindhi) and low of 0.04 (Tamil).

## 4. Experimental Baselines

The three sections of our collection offer several interesting tasks to investigate, and we provide initial baselines for three of them: single word transliteration, full sentence transliteration, and language modeling. These baselines are intended to give some sense of how to approach model validation with this dataset, not necessarily to squeeze every last bit of accuracy on the tasks. Much of the hyper-parameter tuning, for example, model dimensions, number of wordpieces, etc., was performed for each task on some subset of the languages on a separate held aside set, and then used for all languages, rather then performing such tuning on a language-by-language basis – although training stopping criteria were, of course, applied for each task independently.

### 4.1. Evaluation metrics

For two of the three tasks, we evaluate transliteration, which can be evaluated in terms of error rates compared to the reference string. Error rates are the number of substitutions, deletions or insertions within a minimum-error rate alignment of the system output with the reference, per token in the reference. Tokens can be taken as individual Unicode characters (what we call character-error rate, CER) or as whitespace delimited substrings (what we call word-error rate, WER). For single word transliteration, we report both CER and WER; for full sentence transliteration, we report just WER. We present all error rates as percentages and provide mean and standard deviation over 5 trials.

For language modeling results, we are evaluating open-vocabulary models (as motivated in § 4.4.), hence we report bits-per-character (BPC). This is a standard measure (related to perplexity) typically applied to character-level language models. Per sample, it is calculated as the total negative log base 2 probability of the correct output character sequence, divided by the number of characters in the output string. Additionally, since within each language we have parallel native script and romanized corpora that we are training and evaluating on, we can follow Cotterell et al. (2018) and Mielke et al. (2019) in comparing language modeling results across parallel data samples by normalizing with a common factor. We call this bits-per-native-character (BPNC): total negative log base 2 probability divided by the number of characters in the native script strings (rather than the romanized strings).

### 4.2. Single word transliteration

For single word transliteration, we train on the training section of the romanization lexicon for each language and validate on the dev section of that language's lexicon. The test section remains in reserve. For each romanization in the dev set (ignoring the number of attestations), we take the Latin script string as input and evaluate the accuracy of the 1-best output of our models. The training data consists of all attested pairs from the training partition of the lexicon, each pair repeated as many times as it was attested.

#### 4.2.1. Methods

**Pair $n$-gram Models.** Pair $n$-gram models, also known as joint multi-gram models (Bisani and Ney, 2008) are a widely used non-neural modeling method for tasks such as grapheme-to-phoneme conversion, as well as for transliteration (Hellsten et al., 2017). The basic idea is to take a word and its transliteration, such as the Tamil version of the English word "temple" presented earlier (டெம்பிள் temple) and use expectation-maximization to derive a character-by-character alignment, such as the following:

ட:t டெ:e ம:m ்:_ ப:p ி:_ ள:l ்:e

Each symbol consists of an input side and an output side, separated by a colon, where the underscore character ( _ ) represents the empty string. For the above sequence, each character on the input side is from the Tamil script (or _ ) and each character on the output side is from the Latin script (or _ ). Note that diacritics such as virama (◌्) are separate characters in this string, and can align to the empty string or to silent characters in the English orthography.

---

[6] Note that the Wikipedia frontend (as in Hindi) may have a setting to coerce digits into one or another Unicode block, further obscuring original author intent.

|      | Character (CER) and word (WER) error rate percentage mean (std) ||||||
|      | pair 6g || transformer || LSTM ||
| Lang | CER | WER | CER | WER | CER | WER |
| --- | --- | --- | --- | --- | --- | --- |
| bn | 14.2 (.02) | 54.0 (.10) | **13.2** (.07) | 50.6 (.12) | 13.9 (.15) | 54.7 (.45) |
| gu | 12.9 (.04) | 53.5 (.17) | **11.9** (.15) | 50.5 (.53) | 12.6 (.06) | 53.3 (.22) |
| hi | 14.7 (.04) | 53.1 (.06) | **13.4** (.21) | 50.0 (.63) | 13.9 (.10) | 53.0 (.46) |
| kn | 7.2 (.04) | 36.6 (.14) | **6.3** (.12) | 33.8 (.47) | 6.8 (.04) | 37.5 (.22) |
| ml | 10.0 (.07) | 44.7 (.22) | **9.0** (.04) | 41.7 (.15) | 9.2 (.03) | 43.7 (.25) |
| mr | 12.4 (.03) | 51.8 (.12) | **11.6** (.10) | 50.3 (.45) | 12.5 (.08) | 54.6 (.47) |
| pa | 17.9 (.07) | 60.6 (.12) | **17.4** (.33) | 59.1 (.95) | 17.5 (.04) | 59.6 (.28) |
| sd | **20.5** (.06) | 62.9 (.17) | 22.0 (.32) | 66.8 (.67) | 20.6 (.11) | 63.5 (.43) |
| si | **9.1** (.01) | 43.2 (.08) | 9.2 (.10) | 45.3 (.16) | 9.3 (.04) | 45.2 (.14) |
| ta | 9.3 (.08) | 36.8 (.26) | 9.4 (.52) | 34.3 (.78) | **8.4** (.12) | 34.7 (.67) |
| te | 6.9 (.02) | 34.4 (.11) | **6.2** (.11) | 32.4 (.46) | 6.8 (.08) | 34.9 (.37) |
| ur | 20.0 (.07) | 64.3 (.16) | 19.5 (.10) | 63.3 (.24) | **19.4** (.08) | 63.4 (.25) |

Table 2: Single word transliteration (Latin to native script) performance in character-error rate (CER) and word-error rate (WER) for three systems, averaged over 5 trials. Best CER for each language is bolded for ease of reference.

Given such a string of "pair" symbols, we can train an $n$-gram language model, which provides a joint distribution over input/output string relations. Converted into a finite-state transducer, this can be used to find the most likely transliterations for a given input string in either direction (native to Latin or vice versa). See Hellsten et al. (2017) for more details on such approaches. For these experiments, we trained 6-gram models with Witten-Bell (1991) smoothing.

**LSTM Sequence-to-sequence.** This baseline treats transliteration as a standard sequence-to-sequence problem, and applies an encoder-decoder architecture to it, similar to the type used for machine translation in Bahdanau et al. (2014). The architecture consists of a deep bidirectional encoder network[7], itself made up of layers containing a forward LSTM and a backward LSTM, connected to a forward decoder LSTM by an attention mechanism.

Character input was embedded at each input timestep with an embedding size of 512. The encoder consisted of 2 layers. The first was a bidirectional layer, where both forward and backward LSTMs contained 256 hidden units. The second layer was unidirectional, and consisted of a forward LSTM with 256 hidden units. The decoder consisted of a 3-layer LSTM with 128 hidden units. The encoder was connected to the decoder by a Luong-style attention mechanism (Luong et al., 2015). Training was performed via the Adam optimizer (Kingma and Ba, 2014) with a batch size of 1024 and a base learning rate of 0.01. Gradients were clipped at 5.0. Maximum output length was capped at 50 characters. During training, for the encoder dropout between LSTM layers was 0.7, while recurrent connections had dropout set to 0.4; and for the decoder dropout between LSTM layers was 0.25, while recurrent connections had dropout set to 0.6.

---

[7] As described in http://opennmt.net/OpenNMT/training/models/.

**Transformer Sequence-to-Sequence.** This baseline is similar to the LSTM sequence-to-sequence baseline, but swaps out the the LSTM encoders and decoders for a more-powerful Transformer architecture (Vaswani et al., 2017). The goal is to see if the additional model complexity results in worthwhile performance gains.

Following common usage in machine translation, we use sub-word tokens when training our transformer models, here and in § 4.3.. For single word transliteration, where a word is the string to be processed, we simply use single characters as our sub-word tokens. We use the architecture from Chen et al. (2018, Appendix A.2), with model dimension of 128, hidden dimension of 1024, 4 attention heads and 4 transformer layers for both encoder and decoder. Dropout was set uniformly to 0.36 and we use the Adam optimizer. Otherwise, the settings are the same as in Chen et al. (2018).

#### 4.2.2. Results

Table 2 presents CER and WER performance of our three transliteration models across all 12 languages, averaged over 5 trials (reporting means and standard deviations) with the lowest CER result for each language bolded for ease of reference. In fact, all three systems fell within 2% absolute CER performance on all of the languages, with the transformer providing the best performance on 8 of the 12 languages, and the LSTM and pair 6g model best on just 2 of the languages each. Note that one of the languages where the non-neural model performed best was the language with the sparsest training data (sd). Overall, we can observe that the Perso-Arabic scripts proved most difficult, and that the Dravidian languages were among the easiest.

### 4.3. Full sentence transliteration

Full sentence transliteration differs from single word transliteration in that the context of the rest of the sentence can help to resolve ambiguities when determining the intended words. Note that we are validating on manual reference annotations (i.e., the dev set from the romanized sentence collection outlined in § 3.3.), but are not training on any given parallel data. Instead, we have native script full sentence text (the training section of the native script text collection outlined in § 3.1.) along with a romanization lexicon for single words (the training section of the lexicons outlined in § 3.2.). This represents a very common use scenario, as stated earlier, since large-scale parallel data is not generally available. We approach this task in two ways: first, with a noisy channel model; and second, with a sequence-to-sequence model trained on simulated romanizations. We additionally provide single word transliteration baselines, which do not make use of any sentence context, but rather transliterate each word independently.

|   | original: | এরপর clear কমান্ডের মাধ্যমে টার্মিনালে/স্ক্রীনে থাকা সব টেক্সট বা লেখা মুছে ফেলা হবে। |
|---|---|---|
|   | manually romanized: | Arpor clear commender madhome terminal/scriena thaka sob texts ba lekha muche fela hobe. |
| (a) | reference: | এরপর কমান্ডের মাধ্যমে টার্মিনালে স্ক্রীনে থাকা সব টেক্সট বা লেখা মুছে ফেলা হবে |
|   | system output: | এরপর ক্লিয়ার কমান্ডের মাধ্যমে টার্মিনালে স্ক্রীনে থাকা সব টেক্সট বা লেখা মুছে ফেলা হবে |
|   | WER: | 1 insertion, 13 reference words = 7.7 |
| (b) | reference: | এরপর clear কমান্ডের মাধ্যমে টার্মিনালে/স্ক্রীনে থাকা সব টেক্সট বা লেখা মুছে ফেলা হবে। |
|   | system output: | এরপর ক্লিয়ার কমান্ডের মাধ্যমে টার্মিনালে/স্ক্রীনে থাকা সব টেক্সট বা লেখা মুছে ফেলা হবে. |
|   | WER: | 2 substitutions, 13 reference words = 15.4 |

Table 3: The original native script sentence and manual romanization, alongside the reference native script string, hypothetical system output and WER for (a) *whitespace* and (b) *pass-through* methods for evaluating full-sentence transliteration.

### 4.3.1. Full sentence evaluation

Note that the romanization lexicons are exclusively between native script words and possible Latin script romanizations of those words, hence they do not cover any characters falling outside of the Latin alphabet on one side (i.e., a-z) or a subset of the native script Unicode block on the other side, which does not cover things like punctuation or digits. As stated in § 3.3., annotators were instructed to realize the sentence in the Latin script, and to pass any substrings not in the native script through to the resulting romanized sentence. Punctuation and digits in the Latin script are left in the Latin script, and those in the native script (or, e.g., Danda) are typically converted to the Latin script equivalents. Further, some amount of Latin script content does appear in the native script sentences. For example, from the Bengali collection, we have the sentence shown in Table 3, which is discussing the use of the "clear" command in a shell script.

From this example we can see several things. First, the Latin script word "clear" appears in the romanized string with no indication that it was also in the Latin script in the source sentence. (Other English words such as "terminal" and "texts" were written in the native script in the source.) The Latin script slash (/) is left as-is in the romanized version, and the end-of-sentence Danda character (।) is 'transliterated' to a period.

Given that we are not relying on given parallel training data, there are several ways we can evaluate transliteration with these sentences. First, we can elide characters on the romanized side that fall outside of the covered Latin alphabet characters (a-z), after de-casing. We can then transliterate these strings. To evaluate the output, we will distinguish two approaches, which we will call *whitespace*, and *pass-through*. Let $\hat{\mathcal{B}} \subseteq \mathcal{B}$ be the native script characters included in the romanization lexicon. In a *whitespace* evaluation, we will treat any reference character $c \notin \hat{\mathcal{B}}$ as whitespace, and evaluate word-error rate versus those strings. In *pass-through* evaluation, we can reintroduce the out-of-vocabulary (not a-z) characters from the romanized side into the final native script system output, and evaluate word-error rate versus the original reference strings.

To illustrate the difference between these two approaches, let's assume that we achieve perfect transliteration of all the words in the input manually romanized sentence in Table 3.[8] Table 3 presents the reference, system output and resulting WER for both of these evaluation approaches. The inclusion of the Latin script word "clear" in the reference increases the number of reference tokens in pass-through evaluation (b) by one, but the inclusion of slash simultaneously reduces the number of reference tokens by one, hence the final number of reference words is the same for this example in both approaches. In both approaches, the second word is an error – an insertion in (a) or a substitution in (b). However, the final token in the sentence has a mismatch in (b) but not in (a), hence its error rate is double.

Using the pass-through approach to evaluate full sentence transliteration, however, also allows comparison with a full sequence-to-sequence approach, that is trained as follows: For each sentence in the native script training data, automatically romanize substrings that are covered in the romanization lexicon and pass other tokens (e.g., any Latin script strings or punctuation) unchanged to the romanized sentence.[9] A sequence-to-sequence model trained on such pairs can be directly evaluated on the kinds of full sentences that we have in the collection, corresponding to evaluation using the pass-through approach.

In our results, we differentiate between transformer models trained with the tokenization approach taken in whitespace evaluation ('whitespace' transformer) and those with full sequence-to-sequence training ('full string' transformer). See § 4.3.2. for details on these variants. For pass-through evaluation of the output of the 'whitespace' transformer, as with the single word baselines and noisy channel output, we reintroduce out-of-vocabulary characters into the output string, whereas the output of the 'full string' transformer can be directly compared with the reference strings without modification.

Finally, a note about how casing in the input Latin script is handled in these experiments. Since the romanization lexi-cons in the dataset are all de-cased, upper-case Latin characters will only appear in the simulated training data when they occur in the output (native script) strings and are passed through to the input string unchanged, as described above. For this reason, in all trials, Latin characters in the input strings are converted to lower-case prior to input to any of the transliteration systems.

---
[8] For illustration, we will assume that the word "clear" is transliterated by the system as ক্লিয়ার.

[9] For deterministically romanizable characters such as Danda, Arabic full stop or native script digits, we convert to the Latin script equivalents in this romanization.

|      | Whitespace evaluation WER% mean (std) | | | | Pass-through evaluation WER% mean (std) | | | | |
|      | single word | | noisy | whitespace | single word | | noisy | whitespace | full string |
| Lang | pair 6g | transformer | channel | transformer | pair 6g | transformer | channel | transformer | transformer |
|---|---|---|---|---|---|---|---|---|---|
| bn | 35.0 (.11) | 32.5 (0.71) | **18.6** (.02) | 19.7 (0.12) | 39.7 (.09) | 37.6 (0.66) | **25.8** (.01) | 27.5 (0.07) | 33.0 (0.14) |
| gu | 34.4 (.07) | 28.1 (1.37) | **16.2** (.03) | 21.8 (1.36) | 34.9 (.08) | 28.5 (1.37) | **17.0** (.02) | 24.5 (1.38) | 25.4 (1.44) |
| hi | 24.6 (.14) | 25.0 (1.70) | **11.0** (.01) | 15.8 (0.24) | 28.0 (.10) | 28.6 (1.68) | **15.3** (.01) | 21.2 (0.13) | 25.7 (0.32) |
| kn | 23.4 (.21) | 21.0 (0.27) | **17.1** (.03) | 18.3 (0.44) | 24.0 (.20) | 21.6 (0.28) | **18.3** (.02) | 20.9 (0.41) | 21.9 (1.02) |
| ml | 39.4 (.69) | 37.3 (0.31) | 23.5 (.04) | **21.4** (0.27) | 39.1 (.68) | 37.0 (0.31) | 23.7 (.04) | **22.2** (0.19) | 22.9 (0.14) |
| mr | 29.2 (.03) | 28.4 (0.62) | 13.8 (.03) | **13.8** (0.07) | 30.8 (.03) | 30.1 (0.60) | 16.3 (.03) | 16.8 (0.07) | **16.0** (0.20) |
| pa | 38.2 (.35) | 36.1 (1.14) | **16.4** (.02) | 19.3 (0.04) | 40.0 (.36) | 39.2 (1.08) | **21.1** (.02) | 25.1 (0.06) | 29.3 (0.46) |
| sd | 55.3 (.13) | 63.5 (1.38) | **26.1** (.07) | 37.3 (1.20) | 55.9 (.10) | 63.6 (1.17) | **29.6** (.06) | 41.3 (0.84) | 43.1 (1.21) |
| si | 37.0 (.03) | 35.9 (0.96) | **20.3** (.02) | 23.0 (0.77) | 37.5 (.02) | 36.4 (0.93) | **21.2** (.02) | 24.7 (0.71) | 26.0 (1.12) |
| ta | 30.7 (.25) | 31.9 (0.95) | 19.3 (.04) | **18.9** (0.08) | 30.7 (.24) | 31.9 (0.93) | 19.9 (.04) | 20.1 (0.09) | **19.6** (0.05) |
| te | 27.6 (.06) | 26.4 (0.22) | **17.0** (.02) | 18.9 (0.10) | 28.0 (.06) | 26.7 (0.20) | **17.9** (.02) | 21.1 (0.06) | 21.2 (0.18) |
| ur | 33.8 (.08) | 44.5 (3.25) | **12.5** (.08) | 19.3 (0.47) | 38.7 (.08) | 48.2 (3.15) | **18.9** (.08) | 26.2 (0.33) | 29.4 (0.18) |

Table 4: Full sentence transliteration (Latin to native script) word-error rate (WER) percent performance, averaged over 5 trials, using both the whitespace and pass-through evaluation methods presented in § 4.3.1..

#### 4.3.2. Methods

**Single word baselines.** For these results, we use the pair 6g and transformer models presented in the § 4.2. to transliterate each word independently.

**Noisy channel.** We combined the pair 6g transliteration model used for the single word transliteration in § 4.2. with an unpruned Katz-smoothed trigram language model over output (native script) sentences, trained on the training portions of the native script Wikipedia collection presented in § 3.1.. We sped up decoding by first extracting the $k$-best transliterations from each word, then combining with the language model.

**Full sentence transformer.** To derive the parallel training data needed for a sequence-to-sequence model, we first trained a single word transformer transliteration model in the opposite direction, from native to Latin script. Then, for each native script substring in the full sentence Wikipedia collection, we used that model to automatically create a romanized version of each instance by sampling from the 8-best romanizations for that substring. We then produced full romanized sentences in two ways, as detailed earlier: by a) treating substrings outside of the basic Latin alphabet (a-z) as part of whitespace; or b) re-introducing substrings outside of a-z in the native script sentence into the romanized sentence, unaltered apart from a deterministic mapping of native script punctuation and digits to their Latin script equivalents.

From this simulated parallel data, sequence-to-sequence transformer models were trained. As stated earlier, we call the transformer trained on data produced by method (a) 'whitespace' transformers; and that from method (b) 'full string' transformers. For training either model, we follow the same approach for transformer model training outlined in § 4.2., with a few changes to the hyperparameters. Instead of single character sub-word tokens, we make use of a dictionary of 32k "word pieces" (over both input and output vocabularies), using the approach outlined in Schuster and Nakajima (2012). We increased the model dimension to 512, the hidden dimension to 2048, the number of heads to 8, number of layers to 6, and reduced the dropout to 0.1. Otherwise we train as detailed in § 4.2..

#### 4.3.3. Results

The four leftmost columns in Table 4 present mean WER (and standard deviation) over five trials for each language for each of our four methods using whitespace evaluation as detailed in § 4.3.1.. Including sentence context in the model – either through noisy channel or transformer modeling – provides a large reduction in error rate versus single word transliteration. Interestingly, the noisy channel approach provides better performance for 9 of the 12 languages, sometimes substantially better. It seems that the closed vocabulary was not overly constraining for most languages, though the highly inflected Dravidian languages of Malayalam and Tamil were both better modeled by the transformer.

The five rightmost columns in Table 4 present results using the pass-through evaluation method outlined in § 4.3.1. – the same four approaches from the whitespace evaluation columns plus the full string transformer models. The results show similar patterns to the whitespace evaluation, with the same 9 languages performing better with the noisy channel approach than either of the transformers. The full string transformer is better than the whitespace transformer in 2 of the remaining 3 languages, but is (sometimes substantially) worse than the whitespace transformer on the 9 languages where the noisy channel model performs best.

### 4.4. Language modeling

Language modeling of romanized text is also something that can be evaluated with this collection. Note, however, that we do not provide a large repository of full sentence romanized text for training. This is a realistic starting point for de-veloping useful language models, since discovery of romanized text in a language on-line will require an initial model for language identification. To get around this limitation, we can take the kind of simulated romanizations used for the full sentence transliteration training data in § 4.3., and use it to train a language model, while evaluating on our human romanized sentences.[10] We look at several methods of simulation and compare the language modeling performance with modeling of the original native script strings, using bits-per-native-character (BPNC, described in § 4.1.) as the means for comparing across training and validation corpora that encode the same information in different scripts.

|      | native script | romanized |  |  |  |  |  |
|------|---------|-------------|------|-------------|------|--------------|------|
|      |         | Viterbi best |  | Sampled 1x |  | Sampled 10x |  |
| Lang | BPC     | BPC  | BPNC | BPC  | BPNC | BPC  | BPNC |
| bn   | 1.64    | 2.58 | 2.80 | 2.19 | 2.37 | 2.15 | 2.34 |
| gu   | 2.05    | 3.18 | 3.90 | 2.24 | 2.75 | 2.15 | 2.64 |
| hi   | 1.70    | 2.25 | 2.48 | 2.15 | 2.38 | 2.14 | 2.36 |
| kn   | 1.77    | 2.00 | 2.32 | 1.88 | 2.19 | 1.85 | 2.15 |
| ml   | 1.54    | 2.10 | 2.28 | 1.84 | 2.00 | 1.79 | 1.95 |
| mr   | 1.82    | 2.22 | 2.44 | 2.13 | 2.34 | 2.03 | 2.23 |
| pa   | 1.93    | 2.76 | 3.16 | 2.34 | 2.68 | 2.25 | 2.57 |
| sd   | 2.15    | 3.62 | 4.45 | 2.98 | 3.67 | 2.92 | 3.59 |
| si   | 1.95    | 2.44 | 2.91 | 2.11 | 2.52 | 2.00 | 2.39 |
| ta   | 1.46    | 2.24 | 2.44 | 1.90 | 2.07 | 1.88 | 2.04 |
| te   | 1.79    | 2.18 | 2.53 | 1.99 | 2.30 | 1.95 | 2.27 |
| ur   | 1.74    | 2.93 | 3.50 | 2.27 | 2.71 | 2.20 | 2.63 |

Table 5: Language modeling results, presenting bits-per-character (BPC) and bits-per-native-character (BPNC) for each language across four conditions: native script; Viterbi sampled; and k-best sampled over 1 or 10 copies of the corpus.

#### 4.4.1. Methods
**LSTM character-based language models.** For all trials, we train a simple LSTM character-based language model (Sundermeyer et al., 2012) with a character embedding size of 200 and a single hidden layer with 1500 units. We use an SGD optimizer with momentum, batch size of 50 and dropout of 0.2.

**Simulating romanized strings.** For comparison, we use two methods to simulate romanized strings. First, for each word in the native script Wikipedia training data, we use the 6g pair transliteration model[11] described in § 4.2.1. to produce the Viterbi best romanization for that word. Then every instance of that native script word in the corpus is replaced with the romanization. Characters falling outside the coverage of the romanization lexicon are left unchanged in the romanized sentence. Alternatively, instead of taking the Viterbi best romanization at each instance of the word, we sample from the 8-best romanizations, according to the distribution placed by the model over that list after applying softmax. Further, we can create a single romanized version of the training corpus, or continue our sampling procedure over multiple copies of the native script training corpus. Here we evaluate 1 copy (1x) and 10 copies (10x). Finally, Unicode characters that occur less than twice in the training corpus for a particular language and condition (native script or romanized) are replaced with the Unicode replacement character (U+FFFD) in both training and evaluation.

#### 4.4.2. Results
Table 5 presents BPC for each language, both for native script text and romanized text, the latter when trained on Viterbi best or sampled romanizations (1x or 10x) of the training corpus. Sampling is clearly far superior to using Viterbi best, and sampling the corpora 10 times provides a modest improvement over a single copy. Comparing the native script and romanized results via BPNC demonstrates that, unsurprisingly, modeling of the text in the Latin script in these languages is a much harder task than modeling in the native script.

### 5. Conclusion
We have presented the Dakshina dataset, a new Wikipedia-derived resource with data relevant to various use cases involving natural language processing of Latin script data in South Asian languages, including transliteration and language modeling of both native script and romanized sentences. Baseline results using a range of different modeling methods indicate that the various tasks explored are challenging and require further research. It is our hope that this data will spur others in the community to investigate new methods for effectively processing South Asian languages written in the Latin script.

### Acknowledgments
The authors would like to thank Daan van Esch, Renee Klerkx, Jeff Levinson, Vlad Schogol, Shankar Kumar, Alexander Gutkin, Oddur Kjartansson and Savio Lawrence for help with annotation and/or useful discussion on aspects of the project.

---
[10] As noted when training full string transliteration models in § 4.3.1., our simulation methods result in only lower-case Latin script strings, other than where Latin script characters occur in the original strings. For that reason, for the romanized trials in this section, we evaluate language models on de-cased reference strings. We leave for future work incorporation of casing information into the simulation to enable evaluation on cased strings.

[11] These joint models can transliterate in either direction, i.e., from native to Latin script or the reverse.